\documentclass[10pt,journal]{IEEEtran}
\usepackage{amsmath,amsfonts}
\usepackage{amsthm}
\usepackage{array}
\usepackage{textcomp}
\usepackage{stfloats}
\usepackage{url}
\usepackage{verbatim}
\usepackage{graphicx}
\usepackage{amssymb}
\usepackage{pifont}
\usepackage{siunitx}
\usepackage{algorithm,algcompatible}
\usepackage{hyperref}
\usepackage[capitalize]{cleveref}
\usepackage{comment}
\usepackage{multirow}
\usepackage{tikz}
\usepackage{pgfplots}
\pgfplotsset{compat=newest}
\usepackage{caption}
\usepackage{tabularx}
\usepackage{xspace}
\usepackage{afterpage}
\usepackage[tight]{subfigure}

\newlength\figH
\newlength\figW
\newlength\vertdis

\usetikzlibrary{shapes}
\usetikzlibrary{circuits.ee.IEC}
\usetikzlibrary{quotes}
\usetikzlibrary{angles}
\usetikzlibrary{positioning}
\usetikzlibrary[pgfplots.groupplots]
\tikzset{RPY/.style={x={(\nxx,\nxy)},y={(\nyx,\nyy)},z={(\nzx,\nzy)}}}

\usepackage{pgfplots}

\newcommand{\code}[1]{\texttt{\small #1}}
\algnewcommand\FUNC{\item[\textbf{Function Signature:}]}
\algnewcommand\INPUT{\item[\textbf{Input:}]}
\algnewcommand\OUTPUT{\item[\textbf{Output:}]}
\algnewcommand\RETURN{\STATE{\textbf{return\ }}}

\definecolor{steelblue48112179}{RGB}{48,112,179}

\DeclareSIUnit{\fps}{fps}

\usepackage{etoolbox}
\usepackage[textsize=scriptsize,textwidth=1.1cm]{todonotes}
  \definecolorseries{test}{rgb}{grad}[rgb]{.95,.55,.55}{11,11,17}
  \resetcolorseries[10]{test}
  \newcommand{\addtodoeditor}[1]{%
    \colorlet{#1}{test!!+!50}
    \expandafter\newcommand\csname#1\endcsname [1]{%
      \todo[color=#1,size=\tiny]{\sffamily\textbf{\uppercase{#1}:} ##1}\xspace%
    }
    \expandafter\newcommand\csname#1i\endcsname [1]{%
      \todo[inline, color=#1]{\sffamily\textbf{\uppercase{#1}:} ##1}\xspace%
    }
  }

  \addtodoeditor{tb}
  \addtodoeditor{az}
  \addtodoeditor{ab}
  \addtodoeditor{fp}
  \addtodoeditor{lw}
  \addtodoeditor{jb}

\usepackage{microtype}

\newcommand{\etal}{\mbox{\emph{et~al.\@}}\xspace}%
\newcommand{\wrt}{\mbox{\emph{w.\@r.\@t.\@}}\xspace}%
\newcommand{\eg}{\mbox{\emph{e.\@g.\@},\@}\xspace}%
\newcommand{\ie}{\mbox{\emph{i.\@e.\@},\@}\xspace}%
\newcommand{\cf}{\mbox{\emph{c.\@f.\@}}\xspace}%

\usepackage[switch]{lineno}

\usepackage{textcomp}
\usepackage{lipsum}
\newcommand\copyrighttext{%
    \footnotesize \textcopyright This work has been submitted to the IEEE for possible publication. Copyright may be transferred without notice, after which this version may no longer be accessible.
}
\newcommand\copyrightnotice{%
    \begin{tikzpicture}[remember picture,overlay]
    \node[anchor=south,yshift=10pt, xshift=10pt] at (current page.south) {\fbox{\parbox{\dimexpr\textwidth-\fboxsep-\fboxrule\relax}{\copyrighttext}}};
    \end{tikzpicture}%
}

\begin{document}

\title{\LARGE \bf A Containerized Microservice Architecture for a ROS 2 Autonomous Driving Software: An End-to-End Latency Evaluation}

\author{Tobias Betz, Long Wen, Fengjunjie Pan, Gemb Kaljavesi, Alexander Zuepke, Andrea Bastoni, Marco Caccamo, Alois Knoll, Johannes Betz

\IEEEauthorblockA{\textit{Technical University of Munich}, Germany}}

\maketitle
\copyrightnotice

\begin{abstract}

The automotive industry is transitioning from traditional ECU-based systems to software-defined vehicles. A central role of this revolution is played by \emph{containers}, lightweight virtualization technologies that enable the flexible consolidation of complex software applications on a common hardware platform.
Despite their widespread adoption, the impact of containerization on fundamental real-time metrics such as end-to-end latency, communication jitter, as well as memory and CPU utilization has remained virtually unexplored. This paper presents a microservice architecture for a real-world autonomous driving application where containers isolate each service. Our comprehensive evaluation shows the benefits in terms of end-to-end latency
of such a solution even over standard bare-Linux deployments.
Specifically, in the case of the presented microservice architecture, the mean end-to-end latency can be improved by \num{5}-\num{8}\%. Also, the maximum latencies were significantly reduced using container deployment.
\end{abstract}


\section{Introduction}
The automotive market is shifting towards software-defined vehicles (SDV), enabling a more software-centric automotive ecosystem.
For example, the open-source consortium SOAFEE \cite{SOAFEE, Spencer2022} specifically targets SDV and brings together OEMs, Tier~1s, and chip manufacturers to work on the challenges.
The E/E architecture of SDVs is based on a central computing unit in which a powerful high-performance computer manages and coordinates diverse functionalities. These functions encompass processing of sensor data, operation of infotainment systems, advanced driver assistance systems, and communication with external systems.
This enables the separation of software from hardware functionality to achieve greater modularity and scalability. Lightweight virtualization techniques such as containerization enable efficient resource utilization and isolation of software components. This design philosophy empowers applications and services to operate independently within their dedicated virtual environments.
When employing virtualization technologies in SDVs, stringent real-time criteria such as latencies and deadlines must be met.
This concern becomes particularly important in autonomous vehicles, where the timely processing of sensor data within a predefined time window is critical for enabling prompt decision-making and control. Typically, end-to-end latencies of \SI{100}{\milli\second} are considered acceptable, wherein the
sensor data must be swiftly processed, and the resulting output variables must be made available from the vehicle's trajectory controller \cite{Lin.03192018}.
Therefore, the end-to-end latency directly impacts the vehicle's ability to navigate and respond to dynamic road conditions in a safe and reliable manner. Failure to meet these real-time requirements could lead to performance degradation and an increased risk of accidents \cite{Betz2022IAC}. In the context of software-defined autonomous driving architectures, practitioners have been experimenting with frameworks that simplify the difficult tasks of configuring, tuning, and optimizing the complex chains of architectural interdependencies. In particular, Autoware~\cite{TheAutowareFoundation.} and the Robot Operating System~(ROS)~2~\cite{ROS2.2022} are among the most widely used frameworks.\\
\begin{figure}
\centering
  \includegraphics[width=\columnwidth]{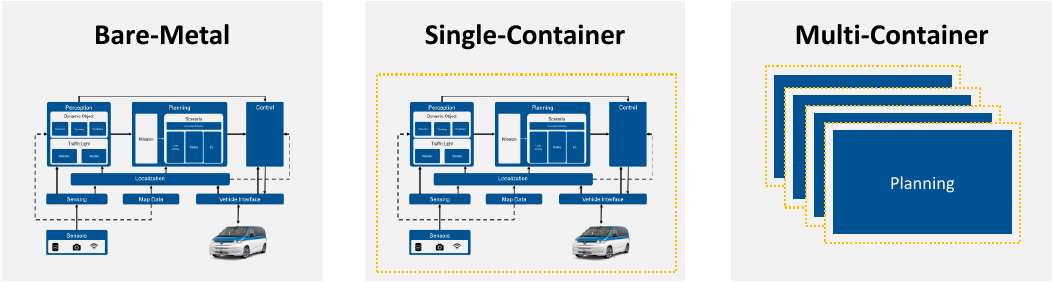}
  \caption{Considered software deployments: bare-metal (no containers), one container, isolation %
  via dedicated containers.}
  \label{Autoware_benchmark}
\end{figure}
This paper introduces a microservice architecture---developed and applied to the research vehicle EDGAR~\cite{karle2023edgar}---that is explicitly designed for Autoware, an open-source autonomous driving software built on ROS~2.
Since the impact of lightweight virtualization technologies on the latency of complex software has---to our knowledge---not yet been considered, this paper investigates the impact of containerization on the end-to-end latency in autonomous driving systems.
Specifically, we focus on the end-to-end latency of a real-world autonomous driving architecture based on Autoware. We deploy the architecture on two different platforms (\code{x86} and \code{aarch64}) using multiple configurations corresponding to an increasing level of container-based isolation (see \cref{Autoware_benchmark}).
Using standard practice in industry \cite{arundel2019cloud}, we use the container orchestration tool \emph{k3s}~\cite{k3s} and Docker \cite{merkel2014docker} to manage the functional dependencies among software packages and containers. Overall, the paper makes the following contributions:

\begin{itemize}
    \item We present the structure and building process of a microservice architecture for autonomous driving software serving as a testbed for future work.
    \item We perform a comprehensive analysis of the impact of containerization using both specific benchmarks and direct measures on increasingly isolated microservice configurations.
    \item We
    quantitatively evaluate
    multiple real-time metrics, including end-to-end latency, jitter, system CPU and memory utilization.
\end{itemize}

Contrary to the common belief, our results show that containers can achieve lower end-to-end latency and better system utilization than bare Linux configurations. This underlines the challenge of finding the best-suited configuration options in very complex system scenarios and shows the benefit of containerization for future SDV systems. The developed microservice architecture will be contributed open-source to the Autoware Foundation (\url{https://github.com/autowarefoundation/autoware}).

\section{Related Work}

Several papers discuss challenges and advancements in embedded systems and automotive software. Sax \etal~\cite{sax2017survey} emphasize the shorter release cycles, increased variants, and software updates in modern vehicles. However, they do not provide any in-depth analysis of particular solutions or tools.
The integration of new functionalities increases the complexity of vehicle systems, requiring careful considerations of the architecture and distribution of electronic control units to effectively manage this complexity~\cite{haas2016cross}.
Kugele \etal~\cite{Kugele2018ElasticSP} discuss elastic service provisioning in intelligent vehicles. The management of different workloads, resource constraints, and changing user requirements is highlighted as a need. Hence, the importance of providing scalable and flexible services that are able to dynamically allocate resources and change performance characteristics based on real-time conditions.

Microservices and service-oriented architectures (SOA) have the potential to improve the flexibility of automotive systems.
Lotz \etal~\cite{Lotz2019} investigate the feasibility and impact of implementing a microservice architecture for driver assistance systems and demonstrate the reduction of complexity and improvement of software systems. Tamanaka \etal~\cite{Bernardino2022} present a conceptual framework for a fault-tolerant architecture and highlight the use of microservices and containerization as critical components. In \cite{Brogi2019DesignPA}, a literature review explores design principles and architectural refinement strategies for microservices. Through a systematic mapping study, Kukulicic \etal~\cite{Kukulicic2022} analyzes the adoption of SOA in automotive software. Functional usability stands out as the most relevant benefit, while issues such as \eg security, safety, and reliability are identified as challenges. Previous research provides overviews, on a purely theoretical basis, of the challenges and benefits of moving to a microservice architecture (see \cite{Velepucha2021}).

Regarding the performance impact of virtualization and containerization technologies on diverse systems,
in \cite{giallorenzo2021virtualization} the authors introduce a benchmarking suite to assess the resource costs of various virtualization technologies. They compare the performance of hardware native hypervisors, hosted hypervisors, and containers using reference benchmarks.
Morabito \etal~\cite{morabito2017virtualization} focus on evaluating the performance of containerization on Internet-of-Things edge environments. The strengths and weaknesses of various low-power devices when dealing with container-virtualized instances are highlighted. Notably, both demonstrate that virtualized or containerized systems show acceptable performance compared to bare-metal systems.
Felter \etal~\cite{felter2015updated} compare the performance of virtual machines (VMs) and Linux containers within cloud computing environments. The study demonstrates that containers outperform or match the performance of VMs in most cases, emphasizing the potential benefits of using containers in cloud architectures.
Similarly, in \cite{xavier2013performance}, the authors conduct research on container-based virtualization in high-performance computing, highlighting the low overhead and potential for near-native performance. While all these studies provide valuable insights, they lack experiments on real-world use cases. In the automotive area, Rajan \etal~\cite{rajan2018hypervisor} explore the technique of bringing virtualization into automotive multicore controllers. The authors evaluate the performance of a virtualized system in terms of core loading, interrupt timing, and task timing parameters.
Long \etal~\cite{Wen2023} develop a general benchmark that yields results consistent with the conclusions mentioned previously. Furthermore, they specifically focus on the startup time of microservice-based Autoware automotive applications demonstrating that virtualization and containerization are suitable and viable options.
The adoption of these virtualization technologies might be beneficial in the automotive industry.
For the development of robot software, ROS~2 is the most widespread framework. \cite{Reke2020} presents an exemplary architecture tailored to autonomous driving and the possibilities of using it for high-speed autonomous racing are presented in \cite{betz2023tum}. The proposed racing architecture is based on microservices where each functional module, \eg perception, planning, and control, is deployed as a container. Autoware \cite{TheAutowareFoundation.} represents the most comprehensive open-source initiative dedicated to ROS~2 for autonomous driving software. In the literature, there are already several different frameworks \cite{li2022autoware_perf, CARET, DAG_Betz.2023, Blass.2021} that allow to measure ROS~2 applications. These are normally based on \code{ros2\_tracing} \cite{Bedard.2022}, which instruments trace points into the middleware accordingly. This allows determining callback times as well as end-to-end latencies. In addition, there are benchmark tools that are mostly limited to simple examples where statements can be made about the DDS latency \cite{Apex.2021, iRobot.2021} or the entire system performance \cite{NvidiaISAAC}. For a single-threaded executor system, timing analysis is performed in \cite{teper2022end}. Reke \etal~\cite{Reke2020} conduct an end-to-end latency and corresponding jitter analysis for an entire application. In \cite{Kato.2018} an analysis for Autoware is realized, with a focus on embedded hardware. The authors in \cite{betz2023latency} analyze the impact of the different system abstraction layers on the end-to-end latency of Autoware. They also tried different Linux scheduling configurations to improve the timing behavior on bare-metal systems. In \cite{Kronauer.2021, Oops2021} the focus is only on the influence of the DDS layer. For the ROS~2 autonomous racing software presented in \cite{betz2023tum}, a latency evaluation is carried out in \cite{Betz2022IAC}. The focus is on the application layer and vehicle stability impact of timing. However, the influence of the microservice architecture is not evaluated. Based on the current state of the art, it is impossible to find an assessment of the impact of containerization and a corresponding microservice architecture on the end-to-end latency of ROS~2 applications.

\section{Microservice Architecture for an Autonomous Driving Software} \label{sec:microservice}

\begin{figure}
\centering
  \includegraphics[width=1.0\columnwidth]{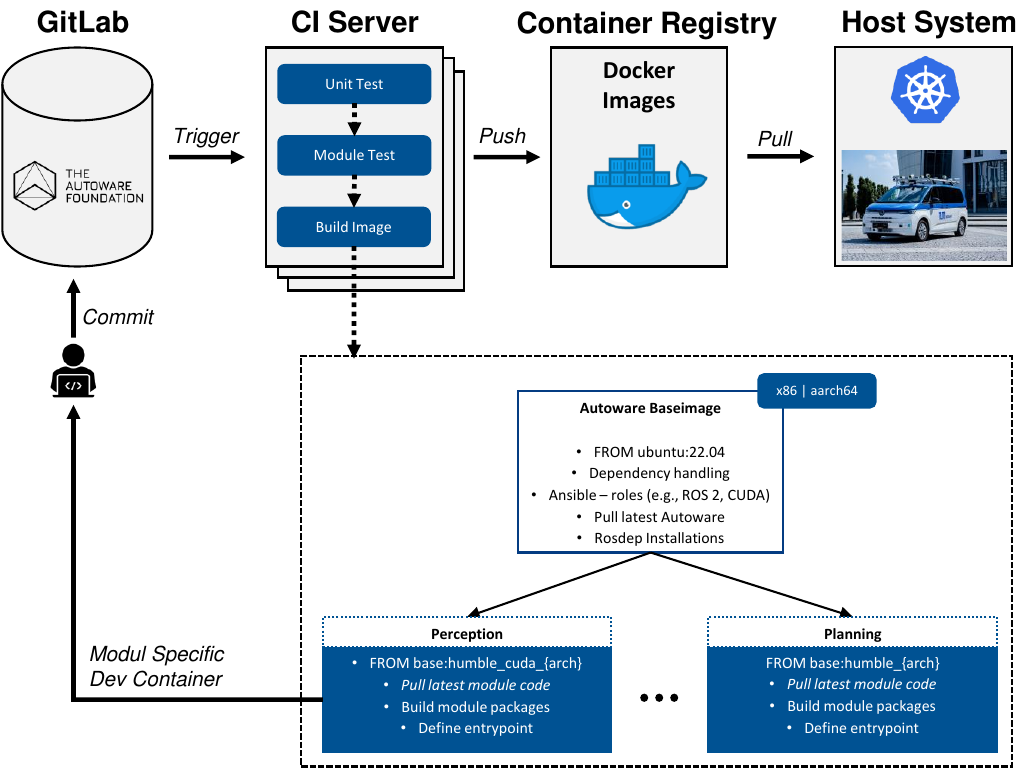}
  \caption{Schematic of the build and deployment process of the microservice architecture: After committing code changes to the Autoware repository on the CI server, the test procedure and docker image build steps are triggered. The built images are stored in the container registry and can be pulled from the cloud onto the host system. The corresponding module images are based on a base image that contains the necessary basic installations. This module image, in turn, also serves as a container for the development of features.}
  \label{fig:build_microservices}
\end{figure}

Following the paradigms of software-definedness, the microservice architecture for autonomous driving software is
designed to enhance the modularity of the software, enabling efficient development and corresponding software deployment.
The core of our architecture is a \emph{base image} that forms the basic building block for the individual module containers. Specialized containers implementing dedicated functionalities are derived from the base image. The base image can also be used for development as it has the requirements for building the complete code. The base image includes essential installations such as ROS~2 and optional libraries like \code{cuda}, \code{cuDNN}, and \code{TensorRT},
which are not necessarily required by every specialized module. The advantages of using a single base image are manifolds. The configuration and installation of all packages can be centralized using a multi-step process that relies on, \eg \emph{Ansible} roles, \emph{rosdep} installation, and manual configuration. This also simplifies the management of cross-package dependencies, facilitates \emph{freezing} packages to specific versions, and avoids introducing incompatibilities between (updated) packages and our code.\footnote{Managing dependencies in ROS~2 is particularly complex and manual optimization, as well as package updates, quickly become a daunting task.}
Once configured, the base image rarely needs to be rebuilt. \cref{fig:build_microservices} depicts the build and deployment process of the microservice architecture.
We divided the Autoware software into eight dedicated containers based on the functional modules in the software. The containers are sensing, perception, localization, map, planning, control, vehicle, and system. Each container consists of multiple ROS~2 nodes, as shown in \cref{tab:service_description}.
In our architecture, the entire ROS~2 launch structure of Autoware was restructured with the separation of individual modules.
The centralized launch package, which listed all packages as dependencies, was split into individual launch packages for each module (with only the needed dependencies).
As a result, each module can be built and launched individually.
The former central launch package included all launch parameters. In contrast, in our architecture, a separate package was created to contain these launch parameters, which are accessible by the module launch files. Additionally, we integrated the launch parameters to be located outside the containers and mounted during the startup of the respective containers.
This approach provides the advantage that changes affecting several module containers, such as the vehicle model, only need to be modified in one location, ensuring consistent parameters for all modules.
We developed a continuous integration (CI) pipeline for building custom module containers that ensures compatibility with both \code{x86} and \code{aarch64} architectures by using cloud-native hardware resources. The CI pipeline consists of several stages that enable both the creation of the entire software and the targeted creation of individual modules. This approach offers efficiency advantages, eliminating the need to rebuild all containers for each code change. Additionally, it facilitates selective updates and maintenance via a CI pipeline-based multi-stage testing process.
Initially, unit tests are conducted, followed by modular tests in which several functions and their interactions are assessed. Due to the modular container structure, a test does not have to be executed repeatedly, but only within the respective container module. We utilize the CI cloud infrastructure to store our built containers in the container registry. The built containers can be seamlessly deployed on both simulation infrastructure and actual vehicles, offering a flexible deployment strategy.
Compared to a monolithic architecture, our microservice architecture improves the development and deployment of the software. During development, the software developer only needs to handle the dependencies related to the respective functionality. The building of the software is automatized in the cloud, and the deployment is simplified.
This development and deployment workflow of the microservice architecture is successfully used in real vehicle projects \cite{karle2023edgar, betz2023tum}.

\begin{table}[] %
\caption{Description of the individual services and number of executed ROS~2 nodes for the Autoware microservices.}
\label{tab:service_description}
\fontsize{7pt}{7pt}\selectfont
\centering
\begin{tabularx}{\columnwidth}{|l|c|X|}
\hline
\textbf{Service} & \textbf{Nodes} & \textbf{Description} \\
\hline
Sensing & 48 & Collecting and pre-processing of raw sensor data \\
\hline
Perception & 49 & Object detection, tracking, and prediction of traffic participants \\
\hline
Localization & 33 & Estimation of vehicle pose, velocity, and acceleration \\
\hline
Map & 6 & Broadcast semantic and geometric information about the environment \\
\hline
Planning & 25 & Generation of the trajectory of the ego vehicle \\
\hline
Control & 8 & Generate control commands to the vehicle \\
\hline
Vehicle & 1 & Passes control signals to the vehicle and receives vehicle information \\
\hline
System & 21 & Error monitoring \\
\hline
\end{tabularx}
\end{table}

\section{Experiments}

\begin{figure}
\centering
  \includegraphics[width=1.0\columnwidth]{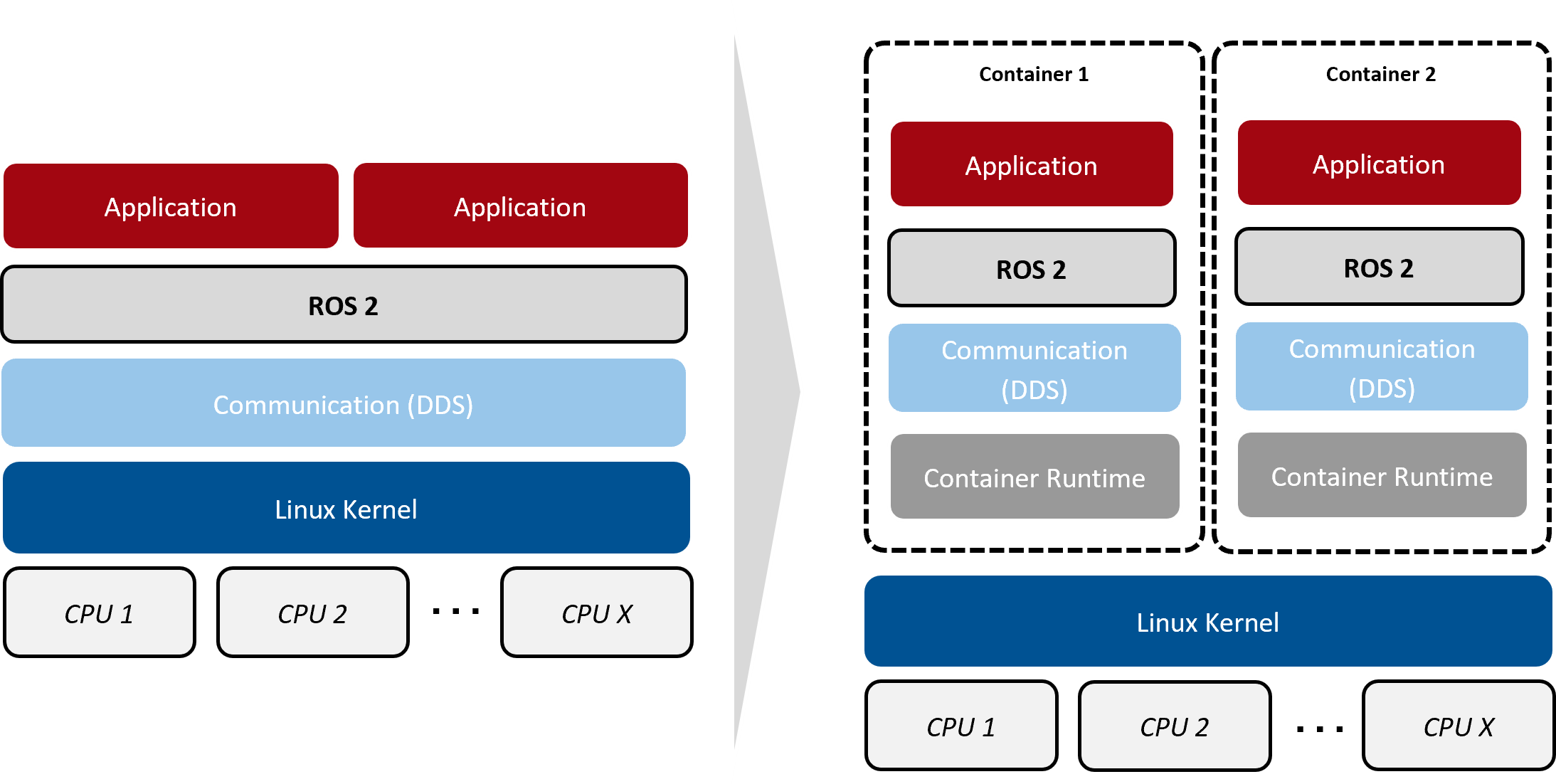}
  \caption{The abstraction layers of a ROS~2 application executed in bare-metal (left) and in multiple containers (right). Using containerization, an additional layer is introduced. Each container consists of independent individual layers, but shares a common Linux kernel.}
  \label{fig:abstraction_layer_experiment}
\end{figure}

A typical ROS~2 application can be abstracted in several layers ranging from high-level applications to the foundational hardware. We define the layers as depicted in \cref{fig:abstraction_layer_experiment}. With the use of containerization, the container runtime adds an additional layer. Positioned above the operating system, this layer facilitates the creation, execution, and management of containers. These executable software packages encapsulate an application and its dependencies.
Our study focuses on understanding the influence of containerization on ROS~2 applications. Specifically, our experiments were systematically designed with an increasing complexity:

\begin{itemize}
    \item DDS Communication: This experiment examines the pure communication performance of DDS in isolation.
    \item ROS~2:  A publish/subscribe example is introduced to observe the performance implications of DDS and ROS~2.
    \item Real-World Autonomous Driving Application:  Incorporates the impact of containerization on the developed microservice architecture.
\end{itemize}
\noindent
For each of the experiments, the three increasing-isolation deployments (Fig.~\ref{Autoware_benchmark}) have been evaluated.
The first scenario (bare-metal) serves as reference point and tests run natively on the system without containerization.
In the second scenario we ran the test within a single container. This aims to measure the overheads introduced by containerizations in the first place.
The third scenario, multi-container, placed the respective benchmark algorithms in separate containers.

In this section, we first introduce our hardware setup and the specific configurations of our containerization architecture. Afterward, we describe the DDS, ROS~2, and Autoware benchmark setups with their individual metrics.

\subsection{Hardware Setup and Software Configurations}

\begin{table}[] %
\renewcommand{\tabularxcolumn}[1]{>{\centering\arraybackslash}p{#1}}
\caption{Specifications of the computing platforms.}\label{tab:hardware_description}
\fontsize{7pt}{7pt}\selectfont
\centering
\begin{tabularx}{\columnwidth}{|l|X|X|}
\hline
 & \textbf{InoNet } & \textbf{ADLINK} \\
 & \textbf{Mayflower-B17} & \textbf{AVA COM-HPC} \\\hline
\rule{0pt}{7pt}\textbf{CPU} & \parbox{0.3\columnwidth}{\centering AMD EPYC 7313P \\ (x86)} & \parbox{0.3\columnwidth}{\centering Ampere Altra Q32-17 \\ (aarch64)} \\ \hline
\rule{0pt}{7pt}\textbf{Clock Frequency} & \parbox{0.3\columnwidth}{\centering \num{16} x \SI{3.0}{\giga\hertz} \\ (max. \SI{3.7}{\giga\hertz})} & \parbox{0.3\columnwidth}{\centering \num{32} x \SI{1.5}{\giga\hertz} \\ (max. \SI{1.7}{\giga\hertz})} \\ \hline
\rule{0pt}{7pt}\textbf{RAM} & \num{4} x \SI{32}{\giga\byte} & \SI{32}{\giga\byte} \\ \hline
\rule{0pt}{7pt}\textbf{GPU}  & \multicolumn{2}{c|}{NVIDIA RTX A6000 \SI{48}{\giga\byte}} \\ \hline
\rule{0pt}{7pt}\textbf{Disk} & \multicolumn{2}{c|}{Samsung 980 PRO NVMe M.2 SSD - \SI{2}{\tera\byte}} \\ \hline
\rule{0pt}{7pt}\textbf{Kernel}  & \multicolumn{2}{c|}{6.2.0-34-generic} \\ \hline
\rule{0pt}{7pt}\textbf{OS}  & \multicolumn{2}{c|}{Ubuntu 22.04.3 LTS Jammy Jellyfish} \\ \hline
\end{tabularx}
\end{table}

All experiments were conducted on two distinct computing platforms  (one \code{x86} and one \code{aarch64} (Armv8)), as depicted in \cref{tab:hardware_description}. The two platforms are representative of autonomous driving platforms for SDVs~\cite{adlink.soafee} and use the same GPU, OS, and Kernel version. On the \code{x86} computing system, we disabled hyperthreading to minimize potential performance fluctuations. The corresponding experiments are performed with ROS~2 \emph{Humble Hawksbill} with the underlying middleware Eclipse CycloneDDS \cite{cyclonedds}. We chose Docker (version 24.0.5) as containerization technology due to its advanced GPU integration capabilities, which provide an obvious advantage over alternative solutions like Podman. To orchestrate the microservice architecture, we utilized k3s (version v1.27.3+k3s1) to deploy and manage the containers. We employed the \code{nvidia-docker2} package to enable GPU support for Docker and the \code{nvidia-device-plugin} for k3s. The container \emph{pods} are configured in such a way that they communicate over the local host network. No CPU requests or limits are set in the configuration. The standard Linux Completely Fair Scheduler (CFS) is used for every experiment.
Despite not being a ``true'' real-time setup, we are interested in replicating a soft real-time environment that reflects the typical setups for software-defined architectures adopted by the practitioners~\cite{SOAFEE, Indychallenge}.

\subsection{Benchmarks}

\subsubsection{DDS Communication} \label{sec: dds}
To test DDS communication, we use the \code{ddsperf} benchmark from the Eclipse CycloneDDS. This benchmark focuses purely on DDS communication, as it skips the ROS~2 abstraction layer. This approach enables us to investigate the influence of containers on pure DDS communication.
The experiment uses a straightforward ``ping pong'' communication pattern to analyze containerization's impact on DDS performance. This pattern consists of continuously sending a defined message size back and forth between two nodes. In the multi-container scenario, each node is placed in an individual container.
CycloneDDS can be configured in two modes: reliable and best-effort.
In the best-effort setting, a publisher sends messages without any assurance that the recipient will receive them correctly. Conversely, in reliable mode, the publisher continues sending messages until it receives an acknowledgment from the subscriber indicating successful reception. Given that best-effort is the default setting for most nodes in the Autoware software, we opt for this mode for our study.
Another crucial aspect was the variation in message size. Starting at \SI{1}{\kilo\byte}, the size was gradually increased by doubling message sizes to analyze the impact on performance across a spectrum of message sizes up to \SI{8}{\mega\byte}. This variation allowed us to assess the scalability and efficiency of DDS communication under different load conditions. Finally, each test was run three times to ensure reproducibility and consistency of results. Each run with a different message size lasted \num{30} minutes. This time period was chosen, in particular, to ensure that a sufficient number of packets could still be exchanged during the tests with the largest message sizes.

\subsubsection{ROS~2} \label{sec:ros}
We used the NVIDIA-ISAAC-ROS \code{ros2\_benchmark} from \cite{NvidiaISAAC} to evaluate the impact of containerization on simple ROS~2 applications.
This benchmark framework is sophisticated and allows testing several example ROS~2 graphs. From the \code{ros2\_benchmark}, we chose the \emph{AprilTag} \cite{Wang2016} node as a reference for our evaluation.
The benchmark includes a playback node that sends camera data, which is in turn processed by the AprilTag detection node. The benchmark also comprises a data-loader node that loads the \emph{rosbag} r2b\_storage data into a buffer and sends it to the playback node. A monitoring node for benchmark-internal evaluations (\eg CPU monitoring) is also included.
In the bare-metal configuration, we run the entire framework without changes to the systems. In the single-container configuration, we put the playback and detector nodes inside the single container, whereas in the multi-container configuration, we separate both nodes into individual containers.
We let the benchmark complete a total of \num{100} runs per each deployment type. Each individual run consists of \num{5} internal iterations. Eventually, the benchmark outputs a statistical result for the five iterations, which we merge accordingly for the \num{100} runs.
In our experiments, the benchmark tests four different setups in terms of the publishing frequency of the playback node: \SI{10}{\fps} (\SI{100}{\milli\second}), \SI{30}{\fps} (\SI{33.3}{\milli\second}), \SI{60}{\fps} (\SI{16.7}{\milli\second}), and an additional setup where the system is configured to achieve the maximum throughput.
With increased framerate, the workload for the system also grows. Therefore different stress levels of the system can be evaluated.

\subsection{Real-World Autonomous Driving Application}
We evaluated the performance impact of containerization on Autoware in the microservice architecture presented in~\cref{sec:microservice}. In the bare-metal setup, the Autoware software is created natively on the system and launched accordingly. In the container environments, Autoware is installed inside of one container. The launch command of the bare-metal variant is defined as an entry point in the container and can then be started with k3s. For the microservice architecture, as previously described, each module has its individual launch command defined in the entry point of the container. For all three deployment variants, it is guaranteed that the same software version is compared. We leverage the orchestration framework proposed in~\cite{betz2023latency} to simulate in a closed-loop the deployed Autoware variants using the \emph{AWSIM} environment. The Autoware software is executed standalone on the described compute platforms, and the simulation is executed on a different compute unit. The vehicle is driving on a defined test route in Nishi-Shinjuku in Tokyo, Japan. Traffic participants were removed from the simulation because they cannot be simulated in a reproducible manner.
Each experiment is repeated until 100 valid runs can be evaluated.
Each test drive takes approximately two minutes to reach the goal pose.

\subsection{Metrics}\label{sec:metrics}

It is important to develop metrics at both application and system levels to analyze the impact of containerization. Such metrics provide valuable insights into resource utilization, helping to identify the latency impact induced by containerization. However, benchmarks are often published with their metrics, making it difficult to evaluate all experiments consistently. In the following, we will go into more detail about the metric used for each experiment.

\subsubsection{DDS Communication}
The benchmark provides the throughput of packets sent during the test period. In addition, the round trip latency is displayed, which is the time it takes for a message to be sent from the source node to the destination node and back again. The benchmark does not provide the CPU load during the execution.
After the tests, we calculate the average round trip time and the average throughput.

\subsubsection{ROS~2}
The framework outputs different metrics for each test node. We evaluate the mean end-to-end latency from sending the raw data until the test node generates an output. This metric is calculated internally in the benchmark via tracing points. Also the mean jitter of the corresponding node is measured. Additionally, the framework provides insight into CPU utilization. We evaluate the average CPU utilization over the test runs.

\subsubsection{Real-World Autonomous Driving Application}
The complex ROS~2 Autoware setup is evaluated using the data-age (end-to-end latency) metric, shown \cite{gunzel2023equivalence} to be equivalent to the reaction time. It is the average of path durations with the same sensor input. For this, the framework of \cite{DAG_Betz.2023} is used, which can determine the end-to-end latency for Autoware accordingly. The computation is based on \code{ros2\_tracing}, which places corresponding trace points in the \code{rclcpp} client library of the ROS~2 middleware. To enable tracing while using the containerized architecture of Autoware, it was necessary to mount specific \emph{LTTng} related file information from the host system into each of the containers. Inside the containers \code{ros2\_tracing} must be enabled. For the bare-metal and containerized measurements, the tracing session was executed on the host system.
The framework computes the total end-to-end latency as well as its individual components:

\begin{itemize}
    \item The \textit{idle} latency or intra-node communication latency defines the time between a subscription callback and a timer callback of a ROS~2 node.
    \item The \textit{communication} latency is the time between publishing and receiving a ROS~2 message via a subscription callback. It corresponds approximately to the time needed for the DDS communication.
    \item The \textit{compute} latency describes the time it takes to process the input from a subscription and publish the corresponding output data to the subsequent node.
\end{itemize}

\begin{table}[b!] %
\caption{ROS~2 callback signatures for the evaluated computation chain.}\label{tab:computation_chains}
\centering
\fontsize{7pt}{7pt}\selectfont
\begin{tabular}{|l|}
\hline
\textbf{Computation Chain} \\ \hline
(0) Filter::(PointCloud2,PointIndices)  \\ \hline
(1) NDTScanMatcher::(PointCloud2)  \\ \hline
(2) EKFLocalizer::(PoseWithCovarianceStamped) \\ \hline
(3) EKFLocalizer::()  \\\hline
(4) StopFilter::(Odometry)  \\\hline
(5) BehaviorPathPlannerNode::(Odometry)  \\ \hline
(6) BehaviorPathPlannerNode::()  \\ \hline
(7) BehaviorVelocityPlannerNode::(PathWithLaneId)  \\ \hline
(8) ObstacleAvoidancePlanner::(Path)  \\ \hline
(9) ObstacleVelocityLimiterNode::(Trajectory)  \\ \hline
(10) ObstacleStopPlannerNode::(Trajectory)  \\ \hline
(11) ScenarioSelectorNode::(Trajectory)  \\ \hline
(12) MotionVelocitySmootherNode::(Trajectory)  \\ \hline
(13) PlanningValidator::(Trajectory)  \\ \hline
(14) Controller::(Trajectory)  \\ \hline
(15) Controller::()  \\ \hline
(16) VehicleCmdGate::(AckermannControlCommand)  \\ \hline
\end{tabular}
\end{table}

Since Autoware consists of a large number of individual computational chains, we selected a single chain for evaluating latency. This chain, detailed in \cref{tab:computation_chains}, was chosen to traverse as many containers as possible for a more accurate assessment of their influence. Furthermore, it represents the critical path with the highest latency in the application.
The quality of service setting is configured to ``keep last,'' operating in best-effort mode with a queue length of 1.
To measure the CPU and memory utilization of Autoware, we recorded the process status using Linux \code{ps}. We recorded the information for all processes every \SI{200}{\milli\second}. As we are interested in the influence of the containerized ROS~2 application, the processes are correspondingly filtered after the session to ROS~2, Docker, Kubernetes, and Autoware processes.

\section{Results}

\subsection{DDS Communication}

\begin{table*}[h!]
\fontsize{7pt}{7pt}\selectfont
\centering
\caption{Mean latency and package count for different message sizes for the \code{ddsperf} benchmark.\\
(BM=bare-metal, SC=single-container, MC=multi-container). Min values across configurations are marked in \textbf{bold}.}
\begin{tabular}{|c|lll|lll||lll|lll|}
\hline
\multirow{3}{*}{Message Size}& \multicolumn{6}{c||}{x86} & \multicolumn{6}{c|}{aarch64} \\ \cline{2-13}
 & \multicolumn{3}{c|}{Mean Latency [\SI{}{\micro\second}]} & \multicolumn{3}{c||}{Mean Package Count} & \multicolumn{3}{c|}{Mean Latency [\SI{}{\micro\second}]} & \multicolumn{3}{c|}{Mean Package Count}\\ \cline{2-13}
 & BM & SC & MC & BM & SC & MC & BM & SC & MC & BM & SC & MC \\ \hline \hline
\SI{1}{\kilo\byte} & 9.40 & 8.40 & \textbf{8.29} & 95183875 & 106363549 & \textbf{107825676}  & 35.08 & \textbf{29.22} & 29.76 & 25466930 & \textbf{30566990} & 30019061 \\ \hline%
\SI{4}{\kilo\byte} & 10.67 & \textbf{9.47} & 9.84 & 83878392 & \textbf{94467569} & 90891488  & 37.47 & 33.83 & \textbf{33.40} & 23865466 & 26422565 & \textbf{26762629} \\ \hline%
\SI{8}{\kilo\byte} & 11.56 & \textbf{10.07} & 10.41 & 77502760 & \textbf{88876974} & 85965143  & 43.24 & \textbf{36.87} & 38.12 & 20685487 & \textbf{24257601} & 23468802 \\ \hline%
\SI{16}{\kilo\byte} & 18.31 & 16.56 & \textbf{16.07} & 48958255 & 54174267 & \textbf{55694824}  & 50.95 & \textbf{40.44} & 41.66 & 17573440 & \textbf{22122799} & 21467480 \\ \hline%
\SI{32}{\kilo\byte} & 26.69 & \textbf{23.58} & 23.90 & 33623951 & \textbf{38049383} & 37530124  & 78.29 & 70.62 & \textbf{69.39} & 11445854 & 12690563 & \textbf{12915312} \\ \hline%
\SI{64}{\kilo\byte} & 40.66 & 37.52 & \textbf{37.14} & 22096636 & 23936672 & \textbf{24175705}  & 126.98 & 103.47 & \textbf{103.42} & 7065036 & 8671388 & \textbf{8672852} \\ \hline%
\SI{128}{\kilo\byte} & 68.57 & 68.64 & \textbf{66.15} & 13104554 & 13088043 & \textbf{13578698}  & 219.03 & \textbf{191.06} & 195.60 & 4099368 & \textbf{4698385} & 4589767 \\ \hline%
\SI{256}{\kilo\byte} & 127.44 & \textbf{125.71} & 127.35 & 7051885 & \textbf{7148614} & 7056868  & 421.19 & \textbf{370.59} & 374.44 & 2132447 & \textbf{2423861} & 2399043 \\ \hline%
\SI{512}{\kilo\byte} & 270.62 & \textbf{269.02} & 269.67 & 3380179 & \textbf{3386639} & 3381805  & 1526.63 & 1359.85 & \textbf{993.96} & 649670 & 716643 & \textbf{944632} \\ \hline%
\SI{1}{\mega\byte} & 693.24 & \textbf{639.32} & 707.37 & 1323072 & \textbf{1427427} & 1290395  & 2415.79 & 2033.56 & \textbf{1868.45} & 401169 & 461185 & \textbf{499301} \\ \hline%
\SI{2}{\mega\byte} & \textbf{1201.41} & 1302.54 & 1709.00 & \textbf{761672} & 704386 & 532031  & 5310.44 & 4662.29 & \textbf{3980.55} & 181039 & 205960 & \textbf{242480} \\ \hline%
\SI{4}{\mega\byte} & 2665.02 & \textbf{2655.18} & 3071.55 & 340980 & \textbf{342625} & 294714  & 9279.03 & 12661.69 & \textbf{6581.91} & 102937 & 92376 & \textbf{139366} \\ \hline%
\SI{8}{\mega\byte} & 5338.55 & \textbf{5233.07} & 5532.21 & 169790 & \textbf{172760} & 165345  & 17157.53 & 24752.60 & \textbf{13326.46} & 54806 & 44914 & \textbf{68176} \\ \hline

\end{tabular}
\label{tab:dds}
\end{table*}

The performance benchmark results offer valuable insights into the effects of containerization. As described in Section~\ref{sec: dds}, we used CycloneDDS and measured the round trip latency of specific message sizes and the number of successfully delivered packets. \cref{tab:dds} presents the measured results for both of the compute platforms from message sizes ranging from \SI{1}{\kilo\byte} to \SI{8}{\mega\byte}. On both platforms, container-based deployment achieves lower latencies than  bare-metal deployment in almost all scenarios.
This effect is more evident on \code{aarch64} and particularly pronounced for small message sizes. On \code{aarch64}, bare-metal configurations never perform better than containers, while only \SI{2}{\mega\byte} bare-metal messages achieve a lower latency on \code{x86}.
The latency improvement is more evident on \code{aarch64} than \code{x86}. For example, for \SI{1}{\kilo\byte}, on \code{aarch64} containers achieve around \num{15}\% lower latency (around \num{8}-\num{10}\% on \code{x86}), while for \SI{64}{\kilo\byte}, the improvement is even higher (\num{18}\% vs. \num{8}\%). Multi-containers can consistently perform better than single-containers. This is the case for large message sizes on \code{aarch64}, where multi-container setups always perform better than single-container starting from \SI{512}{\kilo\byte} message sizes. This trend is not confirmed on \code{x86}, where single-container setups perform better for large message sizes.
In a real application such as the Autoware software stack shown later, the observed message sizes are in the range of \SI{1}{\kilo\byte} to \SI{128}{\kilo\byte}. This is a range where containerized versions on both systems showed smaller latencies.

\subsection{ROS~2}

\begin{table*}[h!]
    \fontsize{7pt}{7pt}\selectfont
    \centering
    \caption{Mean latency, jitter, and CPU utilization for different setups of the \code{ros2\_benchmark}.}
    \begin{tabular}{|c|c|lll|lll|lll|}
    \hline
    \multirow{2}{*}{Architecture} & \multirow{2}{*}{Setup}  & \multicolumn{3}{c|}{Mean Latency [\SI{}{\milli\second}]} & \multicolumn{3}{c|}{Mean Jitter [\SI{}{\milli\second}]} & \multicolumn{3}{c|}{Mean CPU Util. [\%]} \\ \cline{3-11}
    & & BM & SC & MC & BM & SC & MC  & BM & SC & MC \\ \hline \hline
    \multirow{4}{*}{x86} & \SI{10}{\fps}& 6.13 & \textbf{5.85} & 5.90 &  2.33 & \textbf{2.09} & 2.14 & 0.96 & \textbf{0.94} & 0.98\\ \cline{2-11}
    & \SI{30}{\fps} & 5.99 & 5.87 & \textbf{5.85} & 1.90 & \textbf{1.83} & 2.07 & 1.60 & \textbf{1.57} & 1.64 \\ \cline{2-11}
    & \SI{60}{\fps} & \textbf{5.76} & 5.80 & 5.81 & 1.62 & \textbf{1.52} & 1.57 & 2.78 & \textbf{2.72} & 2.81 \\ \cline{2-11}
    &  Max. TP & 7.20 & \textbf{7.08}  & 7.26 & 1.03 & \textbf{1.02} & \textbf{1.02}  & 3.59 & \textbf{3.57} & 3.63\\ \hline \hline
    \multirow{4}{*}{aarch64} & \SI{10}{\fps}& 17.82 & \textbf{17.74} & 17.83 &  1.43 & 1.00 & \textbf{0.98} & \textbf{1.11} & 1.13 & 1.15\\ \cline{2-11}
    & \SI{30}{\fps} & 17.65 & \textbf{17.58} & 17.71 & 1.42 & 1.40 & \textbf{1.37}  & \textbf{2.19} & 2.24 & 2.29 \\ \cline{2-11}
    & \SI{60}{\fps} & \textbf{22.76} & 24.73 & 23.98 & 1.23 & 1.19 & \textbf{1.12} & 2.19 & \textbf{2.15} & 2.31 \\ \cline{2-11}
    &  Max. TP & \textbf{18.03} & 18.27  & 18.09 & 1.61 & \textbf{1.59} & 1.62 & 3.57 & \textbf{3.55} & 3.60\\ \hline
    \end{tabular}
    \label{tab:nvidia}
    \end{table*}

We further investigated the combined performance implications of DDS and ROS~2 using \code{ros2\_benchmark}. We present the experimental setup in Section~\ref{sec:ros}. %
Image data with a size of \SI{0.92}{\mega\byte} is transferred for the data set used.
In addition to the pure DDS time, the end-to-end latency now also includes the computation time of the detection algorithm. Therefore,
in percentage terms, the DDS time has a much smaller share. \cref{tab:nvidia} shows the measured results of the conducted benchmark. Contrary to \code{ddsperf}, latency values in \code{ros2\_benchmark} are very close across all setups. Although containerization can achieve slightly lower latency than bare-metal for low \SI{}{\fps} (\num{10} and \num{30}), bare-metal performs slightly better at \SI{60}{\fps}. Differences in latency are minimal (between \num{0.4}\% and \num{4.6}\%). Looking at the jitter shows a reduction due to containerization in almost all cases. For the \SI{10}{\fps} setup, the occurring jitter is reduced by \num{10.3}\% for single-container and \num{8.1}\% for multi-container. At \SI{60}{\fps} by \num{6.2}\% and \num{3.10}\%. At maximum throughput, only minor differences occurred. An outlier occurs in the multi-container deployment only in the \SI{30}{\fps} setup. The \code{aarch64} platform also shows this behavior in all setups except for the maximum throughput. At \SI{10}{\fps}, the jitter is reduced by \num{30.1}\% and \num{31.4}\%. Again, for the other setups, we observe the same behavior.
Only the multi-container deployment in the last setup exhibits slightly increased jitter. Overall, we conclude that containerization may lead to a reduction in latency. Additionally, we observe slight differences in CPU utilization. Specifically, for \code{x86}, the single-container setup demonstrates the lowest utilization compared to other deployments. Conversely, for \code{aarch64} at lower \SI{}{\fps}, the bare metal benchmark outperforms the containerized benchmark, while for higher \SI{}{\fps}, the single-container setup performs the best.

\subsection{Real-World Autonomous Driving Application}
\begin{figure}[t!]
    \centering
\subfigure[Bare-metal setup]{%
    \begin{minipage}[b]{0.24\textwidth}
        \includegraphics[width=\textwidth]{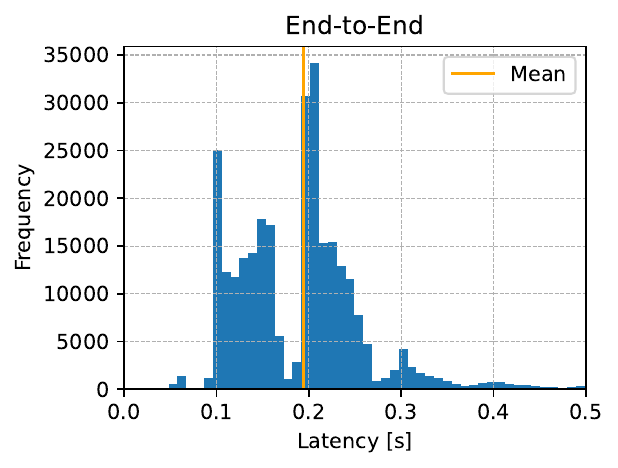}
    \end{minipage}
    \begin{minipage}[b]{0.24\textwidth}
        \includegraphics[width=\textwidth]{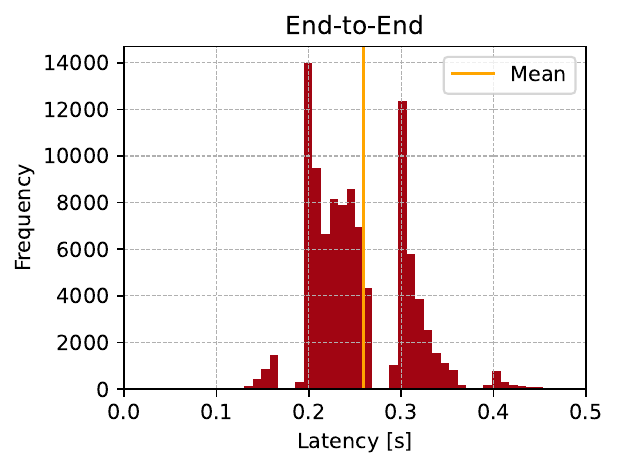}
    \end{minipage}
    \label{fig:bare_metal_experiment}
}%
\hspace{\columnsep}%
\subfigure[Single-container setup]{%
    \begin{minipage}[b]{0.24\textwidth}
        \includegraphics[width=\textwidth]{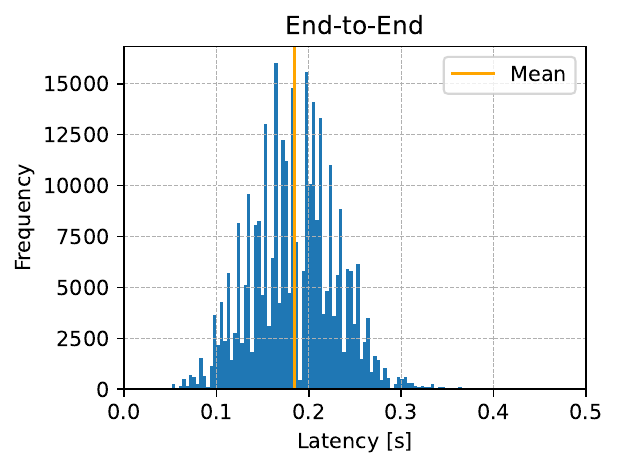}
    \end{minipage}
        \begin{minipage}[b]{0.24\textwidth}
        \includegraphics[width=\textwidth]{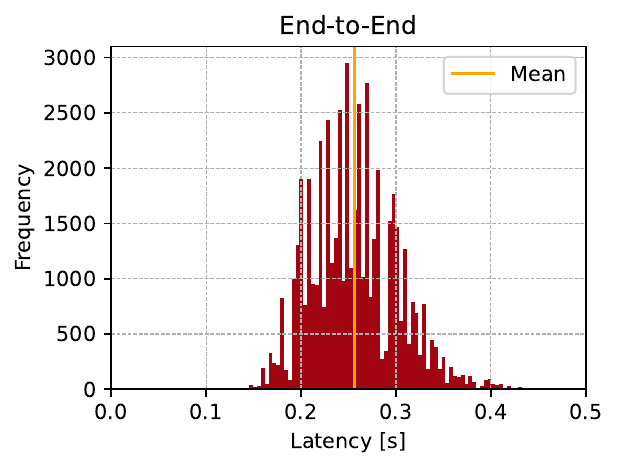}
    \end{minipage}
    \label{fig:single_container_experiment}
}%
\hspace{\columnsep}%
\subfigure[Multi-container setup]{%
    \begin{minipage}[b]{0.24\textwidth}
        \includegraphics[width=\textwidth]{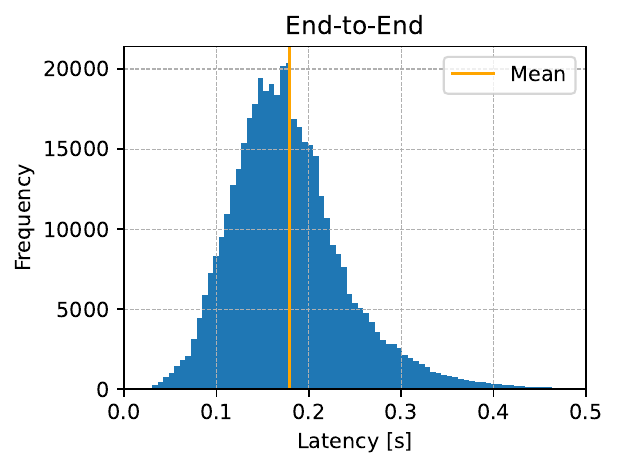}
    \end{minipage}
    \begin{minipage}[b]{0.24\textwidth}
        \includegraphics[width=\textwidth]{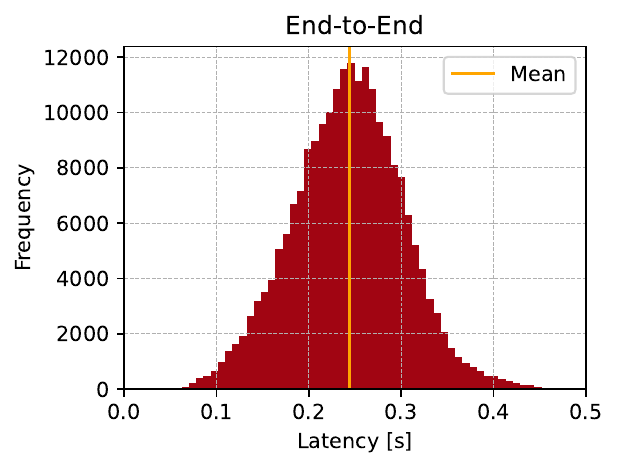}
    \end{minipage}
    \label{fig:multi_container_experiment}
}%
    \caption{End-to-end latency histograms
    (a) using the bare-metal setup,
    (b) using the single-container setup,
    and
    (c) using the multi-container setup (blue=\code{x86}, red=\code{aarch64}).}
    \label{fig:container_experiments}
\end{figure}

\begin{table*}
\fontsize{7pt}{7pt}\selectfont
\centering
\caption{Consolidated statistics for different metrics for Autoware.}
\begin{tabularx}{\textwidth}{| c | c | c | X X X X X X X X X X |}
\hline
\multicolumn{1}{|c|}{Architecture} & \multicolumn{1}{c|}{KPI} & \multicolumn{1}{c|}{Type} & Mean & Std & Skew & Kurtosis & Min & Q25 & Q50 & Q75 & P99 & Max \\
\hline
\hline
\multirow{12}{*}{x86} & \multirow{3}{*}{E2E} & BM & 194.67 & 84.68 & 3.01 & 19.94 & 31.28 & 139.84 & 199.43 & 223.85 & 515.87 & 1437.94 \\
& & SC & 184.51 & \textbf{47.17} & \textbf{0.18} & \textbf{0.35} & 35.37 & 152.59 & 183.86 & 213.95 & \textbf{300.79} & \textbf{553.79} \\
& & MC & \textbf{178.73} & 64.42 & 1.19 & 3.59 & \textbf{25.77} & \textbf{136.00} & \textbf{171.30} & \textbf{210.20} & 383.03 & 903.45 \\
\cline{2-13}
& \multirow{3}{*}{Idle} & BM & 133.81 & 73.12 & 3.04 & 21.88 & 5.24 & 87.09 & 129.82 & 165.77 & 406.713 & 1274.48 \\
& & SC & \textbf{125.92} & \textbf{44.89} & \textbf{0.26} & \textbf{0.44} & 7.33 & 95.85 & 125.71 & \textbf{156.61} & \textbf{237.34} & \textbf{504.89} \\
& & MC & 127.78 & 62.05 & 1.38 & 4.25 & \textbf{3.75} & \textbf{85.78} & \textbf{118.42} & 156.91 & 331.79 & 841.68 \\
\cline{2-13}
& \multirow{3}{*}{Communication} & BM & 8.10 & 10.97 & \textbf{4.83} & \textbf{43.31} & 1.30 & 2.88 & 4.09 & 8.26 & 56.05 & 402.42 \\
& & SC & 3.14 & \textbf{1.79} & 28.52 & 1372.51 & \textbf{1.16} & 2.53 & 2.91 & 3.42 & \textbf{6.59} & \textbf{136.17} \\
& & MC & \textbf{3.12} & 3.20 & 32.91 & 1559.51 & 1.27 & \textbf{2.45} & \textbf{2.90} & \textbf{3.32} & 6.79 & 298.26 \\
\cline{2-13}
& \multirow{3}{*}{Computation} & BM & 52.77 & 24.39 & 3.60 & 56.22 & \textbf{8.56} & 37.68 & \textbf{52.19} & \textbf{61.28} & 124.27 & 623.96 \\
& & SC & 55.46 & \textbf{17.30} & \textbf{0.02} & \textbf{0.91} & 13.41 & 42.75 & 59.74 & 65.80 & \textbf{104.70} & \textbf{175.43} \\
& & MC & \textbf{47.83} & 20.71 & 0.11 & 0.36 & 9.05 & \textbf{32.34} & 53.72 & 62.10 & 105.60 & 244.08 \\
\hline \hline
\multirow{12}{*}{aarch64} & \multirow{3}{*}{E2E} & BM & 258.65 & 71.42 & 3.71 & 28.26 & 120.21 & 210.70 & \textbf{242.14} & 301.10 & 537.49 & 1389.36 \\
& & SC & 256.65 & \textbf{46.41} & \textbf{0.60} & \textbf{0.85} & 144.74 & 221.71 & 251.80 & 284.49 & \textbf{389.62} & \textbf{606.88} \\
& & MC & \textbf{244.11} & 66.08 & 1.39 & 11.44 & \textbf{55.02} & \textbf{202.66} & 243.77 & \textbf{282.81} & 405.38 & 1168.96 \\
\cline{2-13}
& \multirow{3}{*}{Idle} & BM & 112.76 & 54.70 & 2.09 & 14.98 & \textbf{6.80} & 74.95 & 101.85 & 151.29 & 269.51 & 981.29 \\
& & SC & \textbf{97.75} & \textbf{44.85} & \textbf{0.68} & \textbf{0.95} & 7.94 & \textbf{64.44} & \textbf{94.20} & \textbf{126.06} & \textbf{223.01} & \textbf{448.31} \\
& & MC & 117.89 & 56.62 & 3.54 & 35.67 & 7.01 & 83.09 & 112.66 & 146.26 & 258.24 & 1084.05 \\
\cline{2-13}
& \multirow{3}{*}{Communication} & BM & 10.67 & 14.82 & \textbf{10.56} & \textbf{162.57} & 4.69 & 6.49 & \textbf{7.18} & 9.35 & 69.42 & 468.11 \\
& & SC & 8.32 & 4.27 & 13.95 & 279.49 & 5.61 & 6.86 & 7.50 & 8.64 & 18.19 & \textbf{146.23} \\
& & MC & \textbf{7.59} & \textbf{3.19} & 13.08 & 312.38 & \textbf{3.47} & \textbf{6.35} & 7.20 & \textbf{8.08} & \textbf{15.20} & 148.73  \\
\cline{2-13}
& \multirow{3}{*}{Computation} & BM & 135.23 & 27.84 & 6.85 & 73.21 & 90.90 & 123.59 & 133.65 & \textbf{140.63} & 245.06 & 704.96 \\
& & SC & 150.58 & \textbf{14.54} & 1.11 & 7.64 & 107.28 & 141.23 & 151.91 & 158.52 & 199.52 & 324.74 \\
& & MC & \textbf{118.63} & 36.90 & \textbf{-0.61} & \textbf{-0.41} & \textbf{24.66} & \textbf{87.86} & \textbf{131.09} & 147.71 & \textbf{182.57} & \textbf{322.10}  \\
\hline
\end{tabularx}
\label{tab:autowaree2e}
\end{table*}

\setlength{\figH}{5cm}
\setlength{\figW}{1.0\columnwidth}
\begin{figure}[t]
	\centering
	\small
	\input{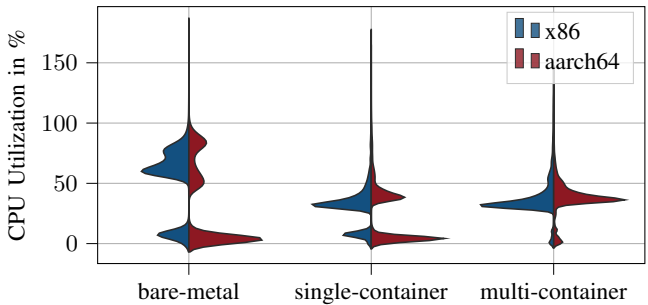}
	\caption{CPU utilization of the systems for the different deployment variants.}
	\label{fig:CPU}
\end{figure}
\begin{figure}[t]
	\centering
	\small
	\input{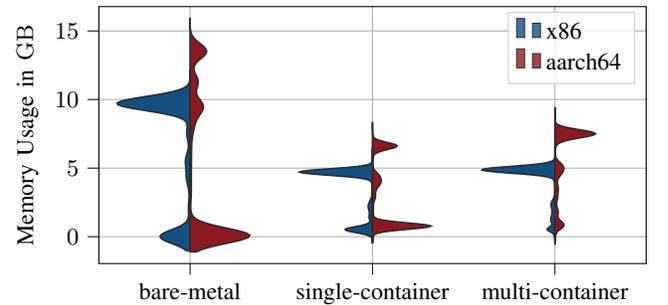}
	\caption{Memory utilization of the systems for the different deployment variants.}
	\label{fig:Memory}
\end{figure}

At last, we evaluate the developed multi-container microservice architecture (see \cref{sec:microservice}) for an autonomous vehicle and compare it with the bare-metal execution and the execution within a single container. We further split the end-to-end latency into its components (Idle, DDS Communication, Computation) to gain a better understanding of each contribution.
Fig.~\ref{fig:container_experiments} shows %
histograms of the different deployment variants on the respective compute platforms. \cref{tab:autowaree2e} presents the detailed measurement values for the entire experiment. As explained in \cref{sec:metrics} the latencies are shown for one computation chain from the sensor to the control output.

\smallskip
\noindent
\textbf{E2E Latency.}
The histograms of bare-metal deployments (\cref{fig:bare_metal_experiment}) show a bimodal distribution with (\code{x86}) high kurtosis value of \num{19.94} and a long tail visible in the Q75 and P99 values (\cref{tab:autowaree2e}).
A similar behavior is evident on the \code{aarch64} system.
Instead, for single-containers, we see a reduced standard deviation and a significantly lower kurtosis of only \num{0.35}. This is reflected in the histogram that show a compact distribution.

The bare-metal implementation of Autoware on the \code{x86} platform shows an end-to-end latency of \SI{194.67}{\milli\second}. Instead, for the developed microservice architecture, the mean latency is reduced by \num{8.1}\% to \SI{178.73}{\milli\second} for \code{x86} and by \num{5.6}\% to \SI{244.11}{\milli\second} for \code{aarch64}.

A relatively large maximum value with \SI{1437.94}{\milli\second} is measured for bare-metal. In the single-container scenario, this maximum value drops to only \SI{553.79}{\milli\second}. We also see a reduction in the various quantiles. From \SI{515.87}{\milli\second} in the 99th percentile to \SI{300.79}{\milli\second}, which is an improvement of \num{41.7}\%.

\smallskip
\noindent
\textbf{Idle Latency.}
Looking at the idle latency, which describes the time data waits for processing via a timer callback, we observe the following for \code{x86}. The native deployment shows an idle latency of \SI{133.81}{\milli\second}. Experiments on \code{aarch64} show lower mean (\SI{112.76}{\milli\second}) and maximum values. Executing Autoware in a single-container environment eliminates the two peaks, resulting in a more even distribution for both \code{x86} and \code{aarch64}. The mean latency value decreases to \SI{125.92}{\milli\second} for \code{x86} and \SI{97.75}{\milli\second} for \code{aarch64}, indicating improvements of \num{5.9}\% and \num{13.3}\%, respectively. This improvement extends to quantile values as well. The 99th percentile sees an improvement of \num{41.6}\% to \SI{237.34}{\milli\second} for \code{x86} and \num{17.3}\% to \SI{223.01}{\milli\second} for \code{aarch64}.
In the multi-container deployment, the mean idle latency is higher than that of single containers, but still lower than the bare-metal setup for \code{x86}. Conversely, a higher measurement value compared to bare-metal is observed for \code{aarch64} (\num{4.5}\%). For \code{x86}, quantiles 25 and 50 are lower compared to single containers, but higher for \code{aarch64}. Both systems exhibit increased values for P99 and the maximum.

\smallskip
\noindent
\textbf{Communication Latency.}
The communication latency represents the smallest portion of the entire end-to-end latency. On the \code{x86} in the native deployment, this latency has a mean of \SI{8.10}{\milli\second}, while on the \code{aarch64}, it has a mean of \SI{10.666}{\milli\second}. However, both systems exhibit maximum values of \SI{402.42}{\milli\second} and \SI{468.110}{\milli\second} respectively, resulting in distributions with long right tails. Notably, the distribution of the single-container variant shows a reduced tail: the mean DDS latency improves by \num{61.2}\% to \SI{3.14}{\milli\second} for \code{x86}, with a smaller improvement (\num{22.0}\%) observed for \code{aarch64}. Both systems also see reductions in their maximum values.
In the multi-container variant, DDS values for both systems are similar to those of the single-container setup, with reduced mean and quantile values within a negligible range compared to single-container. Only the maximum value increases slightly to \SI{298.26}{\milli\second} for \code{x86}, still smaller than in the bare-metal deployment. On the \code{aarch64}, the maximum value is only slightly higher than that of the single-container setup.

\smallskip
\noindent
\textbf{Computation Latency.}
The computation latency is reduced in the multi-container deployment, resulting in improved mean values (\SI{47.83}{\milli\second} for \code{x86} and \SI{118.63}{\milli\second} for \code{aarch64}). These improvements lead to lower values than those observed in both the bare-metal and single-container scenarios. However, we note increased values for P99 and the maximum on the \code{x86} variant.
However, the mean values diverge significantly, and the \code{aarch64} system experiences higher values compared to before. The results deviate from a normal distribution, as no clear peak is visible, but rather multiple peaks in both cases.

\smallskip
\noindent
\textbf{CPU and Memory Utilization.}
\cref{fig:CPU} shows the distributions of CPU usage measured over all runs, \ie, from the execution of the ROS~2 launch file until the vehicle reaches the target position.
We observe the same behavior noted by \cite{Wen2023} regarding the startup of the software.  With bare-metal and single-container, the mean ramp-up phase of the software lasted \SI{13.60}{\second} and \SI{14.4}{\second}, on the \code{x86}. On the \code{aarch64} it is \SI{21.3}{\second} and \SI{23.1}{\second}. Multi-container has a much lower ramp-up phase, where all nodes of the software startup the fastest, \SI{3.8}{\second} for \code{x86} and \SI{4.6}{\second} for \code{aarch64} respectively. This ramp-up phase is clearly visible in both of the plots in the lower part of the graph. As visible in~\cref{fig:CPU} and~\cref{fig:Memory}, containerized applications, regardless of whether single or multi-container, have a lower CPU and memory utilization.
Further analyzing CPU utilization, a significant scatter is observed for bare-metal, ranging from approximately \num{50}\% to \num{90}\% after node initialization. The variance is notably lower for the single-container setup, with a slight difference in utilization, where \code{aarch64} exhibits slightly higher values compared to \code{x86}.
Regarding memory consumption, bare-metal deployments on \code{x86} cluster at around \SI{10}{\giga\byte} RAM, whereas \code{aarch64} displays higher memory consumption with a significantly higher variance. Once again, transitioning the application to a container environment, whether single- or multi-container, leads to a reduction in memory consumption by almost a factor of two for \code{x86}. Similarly, on \code{aarch64}, a drastic decrease in memory consumption is observed. However, both container variants still exhibit higher memory consumption compared to the \code{x86} platform, as seen with bare-metal deployment.

\section{Discussion}
The results of our research uncover unexpected insights into the performance of containerized applications, particularly with respect to end-to-end latency and system utilization. Our results suggest that applications deployed in a container environment have a better latency compared to applications running directly on bare-metal. In the real-world application, end-to-end latency improvements of up to \num{5.2}\% were achieved. The developed microservice architecture showed an improvement of \num{5}-\num{8}\% in the mean. For the maximum values, it was apparent that the single-container had significantly reduced max values %
The DDS Communication benchmark showed that for smaller message sizes, containerization produced better results. This margin was considerably lower (almost absent) in the \code{ros2\_benchmark}. However, in this benchmark, containerization-related jitter was lower than bare-metal.

To better understand the root causes of such behaviors,
we have performed several attempts to optimize the bare-metal Linux system to achieve better results than containers. However, the complexity of the applications considered and their internal interaction is so high that it was not possible to have all parameters under control. We tried to improve the latency with different real-time scheduling algorithms and patches. Nevertheless, the Autoware software starved when we utilized the entire cores for the software. Reserving resources for the Linux processes led to a higher latency compared to the presented results.

At their core, containers leverage kernel parameters and settings to isolate processes using namespaces and cgroups.
Achieving better performance on bare metal typically involves tuning kernel parameters and settings. Isolation is likely a primary factor contributing to the improved performance of containers in our complex setup. By isolating processes, containers ensure that standard Linux processes do not interfere with those inside the container, thereby facilitating an optimized execution environment.

Interestingly, our results showed that deploying applications in multiple containers enhances the improvements of single container configurations, particularly for average end-to-end latencies. This implies that distributing workloads across multiple containers can optimize the overall system performance, particularly in terms of latency. However, this approach can also result in significantly higher maximum execution times. Such trade-offs must be carefully considered when designing and optimizing a system, especially in real-time environments where small maximum execution time is critical.

A critical factor in this discussion is the role of cgroup scheduling and task assignment to cores within the Linux CFS. Platforms like Kubernetes and Docker use this mechanism to effectively schedule container workloads. \emph{Cshares}, a core component of this system, are influenced by various parameters such as predefined CPU limits and the number of processes or threads within a container. The CFS then allocates resources to Cshares, determining how resources are distributed among containers. One of the key advantages of this system is the relative isolation it offers. In a native system, without the protective containerization layer, processes could inadvertently impact each other. For instance, native processes could affect the performance of the Autoware software within the CFS scheduler. Containerization effectively segregates processes, ensuring that each operates within its own domain and remains unaffected by external entities. The inherent mechanisms of cgroup scheduling and its relation with the Linux CFS could play a crucial role in these results.

Another positive effect is that containerization improves latencies by increasing
second-order effects such as the locality of data by grouping related tasks on a smaller set of cores, preventing unregulated migrations to distant cores (our systems have \num{16} and \num{32} cores respectively).
Unregulated migration could be the cause of the bimodal distribution observed for end-to-end latency in \cref{fig:bare_metal_experiment}.
However, a number of experiments and testbeds were implemented to confirm this. This reason could not be confirmed as the sole cause of the performance behavior.
Our study also showed another interesting trend. As the complexity of the test cases increased, the number of processes or threads working within the container also increased significantly. This suggests that as the complexity increases, the container environment becomes more densely populated with processes and threads to handle the increased requirements. 

In summary, our exploration of the containerized applications domain has confirmed the potential benefits of such an approach, not only in terms of isolation but also in terms of performance optimization. This is also observed in the paper \cite{Wen2023}, where investigations of the start-up time of nodes to complete launch coincide with our observation of runtime.

\section{Conclusion}
We presented a microservice architecture tailored to an open-source software for autonomous driving. Our study provided a comprehensive overview of the continuous integration and development process associated with this architecture.
We analyzed multiple metrics for a real-world ROS~2 autonomous driving application based on Autoware and deployed on increasingly isolated container environments.
In order to determine the impact of containerization on communication and simple ROS~2 examples, the analysis was complemented with dedicated benchmarks for DDS and ROS~2.

Our findings indicate that the effect of containerization on runtime varies depending on the complexity of the scenario. In simpler scenarios, the impact of containerization was relatively minor, but, in more complex scenarios, such as that of Autoware, the influence---especially on end-to-end latency---was significant. Moreover, both CPU and memory usage were reduced, leading to improved software stability. These effects were observed and validated on two distinct systems: \code{x86} and \code{aarch64} compute platforms. This cross-system analysis enhances the generalizability of our results.

While our study shows the positive impact of containerization, the complex interactions between containers, Linux CFS, cgroups and Autoware framework require more detailed investigation to determine the exact contributions of each of the mechanisms involved.
However, to our knowledge, this work is the first to provide such in-depth insights into complex real-world autonomous driving setups and highlights the need for more detailed studies in the future.

Looking ahead, there are several directions for further research.
One is to explore strategies to optimize node assignment to containers and the impact of static container allocations to CPUs and setting bounds on CPU shares.
Another interesting topic is hierarchical scheduling, which should be explored in depth to improve the performance of containerized ROS~2 applications.
Furthermore, it is worth considering the generalizability of these results beyond ROS 2 applications.

\label{last-page}

\section*{Acknowledgements}
\noindent T. Betz, as the first author, was the initiator of the research idea and is responsible for the presented concept and implementation.
L. Wen, F. Pan, and A. Knoll contributed to implementation and design of the benchmarks. G. Kaljavesi contributed to the implementation of the microservice architecture. A. Zuepke, A. Bastoni, and M. Caccamo contributed to the evaluation of the performance impacts and the design of experiments.
J. Betz contributed to the conception of the research project and revised the paper critically for important intellectual content. He gave final approval of the version to be published and agrees to all aspects of the work. As a guarantor, he accepts responsibility for the overall integrity of the paper.
M. Caccamo was supported by an Alexander von Humboldt Professorship endowed by the German Federal Ministry of Education and Research.

\bibliographystyle{IEEEtran}
\bibliography{main}
\end{document}